%% file: main.tex
\documentclass[11pt,a4paper]{article}
\pdfoutput=1
\input{commands}
\title{FAME: Feature-Based Adversarial Meta-Embeddings \\for Robust Input Representations}

\author{Lukas Lange$^{1,2,3}$ \\
	\And
	Heike Adel$^1$ \\
	\hspace{4cm}$^1$ Bosch Center for Artificial Intelligence, Renningen, Germany\\
	\hspace{4cm}$^2$ Spoken Language Systems (LSV), Saarland University, Saarbr\"{u}cken, Germany\\
	\hspace{4cm}$^3$ Saarbr\"{u}cken Graduate School of Computer Science, Saarbr\"{u}cken, Germany\\
	{\tt \hspace{4cm}\{Lukas.Lange,Heike.Adel,Jannik.Stroetgen\}@de.bosch.com} \\
	{\tt \hspace{4cm}dietrich.klakow@lsv.uni-saarland.de} \\
	\And
	Jannik Str\"{o}tgen$^1$ \\
	\And
	Dietrich Klakow$^2$ \\
	\\}

\date{}

\begin{document}
	\maketitle
	\input{sections/abstract_fame}

	\input{sections/introduction_fame}
	\input{sections/related_work}
	\input{sections/approach_fame}
	\input{sections/experiments_fame}

\input{sections/analysis_fame}
	\input{sections/conclusions_fame}
	
	\bibliography{refs}
	\bibliographystyle{acl_natbib}

\end{document}

%% file: commands.tex
\usepackage{emnlp2021}
\usepackage{times}
\usepackage{latexsym}

\usepackage{microtype}

\usepackage{amsmath,amssymb}
\usepackage{xspace}
\usepackage{booktabs}
\usepackage{graphicx}
\usepackage{multirow}
\usepackage{subcaption}    
\usepackage{siunitx}

\newcommand{\fastText}{FastText\xspace}

\newcommand{\size}{dimension\xspace}
\newcommand{\sizes}{dimensions\xspace}
\newcommand{\Sizes}{Dimensions\xspace}
\newcommand{\sizesA}{Dim.\xspace}
\newcommand{\types}{granularities\xspace}
\newcommand{\Types}{Granularities\xspace}
\newcommand{\typesA}{Gran.\xspace}

\newcommand{\finetuning}{fine-tuning\xspace}

\newcommand{\finetuned}{fine-tuned\xspace}
\newcommand{\Finetuned}{Fine-tuned\xspace}
\newcommand{\pretraining}{pre-training\xspace}

\newcommand{\pretrained}{pre-trained\xspace}

\newcommand{\subword}{subword\xspace}
\newcommand{\subwords}{subwords\xspace}
\newcommand{\Subword}{Subword\xspace}

\newcommand{\fscore}[1][1]{$F_{#1}$\xspace} 

%% file: sections/abstract_fame.tex
\begin{abstract}
Combining several embeddings typically improves performance in downstream tasks as different embeddings encode different information. It has been shown that even models using embeddings from transformers still benefit from the inclusion of standard word embeddings. 
However, the combination of embeddings of different types and \sizes is challenging.
As an alternative to attention-based meta-embeddings, we propose feature-based adversarial meta-embeddings (FAME) with an attention function that is guided by features reflecting word-specific properties, such as shape and frequency, and show that this is beneficial to handle \subword-based embeddings. 
In addition, FAME uses adversarial training to optimize the mappings of differently-sized embeddings to the same space. 
We demonstrate that FAME works effectively across languages and domains for sequence labeling and sentence classification, in particular in low-resource settings. 
FAME sets the new state of the art for POS tagging in 27 languages, various NER settings and question classification in different domains.
\end{abstract}

%% file: sections/introduction_fame.tex
\section{Introduction}\label{sec:introduction}
Recent work on word embeddings and pre-trained language models has shown the large impact of language representations on natural language processing (NLP) models across tasks and domains \cite{bert/Devlin19,beltagy-etal-2019-scibert,conneau2019unsupervised}.
Nowadays, a large number of different embedding models are available with different characteristics, such as different input \types (word-based \cite[e.g.,][]{mikolov-word2vec,pennington-etal-2014-glove} vs.\ \subword-based \cite[e.g.,][]{bpemb/heinzerling18,bert/Devlin19} vs.\ character-based \cite[e.g.,][]{ner/Lample16,ma-hovy-2016-end,elmo/Peters18}), or different data used for \pretraining (general-world vs.\ specific domain).
Since those characteristics directly influence when embeddings are most effective, combinations of different embedding models are likely to be beneficial \cite{tsuboi-2014-neural,kiela-etal-2018-dynamic,lange-etal-2019-nlnde-meddocan}, even when using already powerful large-scale \pretrained language models \cite{flair/Akbik18,yu2020named}.
Word-based embeddings, for instance, are strong in modeling frequent words while character-based embeddings can model out-of-vocabulary words. Similarly, domain-specific embeddings can capture in-domain words that do not appear in general domains like news text.

Different word representations can be combined using so-called meta-embeddings. There are several methods available, ranging from concatenation \cite[e.g.,][]{yin-schutze-2016-learning}, over averaging \cite[e.g.,][]{coates-bollegala-2018-frustratingly} to attention-based meta-embeddings \cite{kiela-etal-2018-dynamic}. However, they all come with shortcomings: Concatenation leads to high-dimensional input vectors and, as a result, requires additional parameters in the first layer of the neural network. Averaging simply merges all information into one vector, not allowing the network to focus on specific embedding types which might be more effective than others to represent the current word. Attention-based embeddings address this problem by allowing dynamic combinations of embeddings depending on the current input token.
However, the calculation of attention weights requires the model to assess the quality of embeddings for a specific word. This is arguably very challenging when embeddings of different input \types are combined, e.g., \subwords and words. 
Infrequent in-domain tokens, for instance, are hard to detect when using \subword-based embeddings as they can model any token. Moreover, both average and attention-based meta-embeddings require a mapping of all embeddings into the same space which can be challenging for a set of embeddings with different \sizes.

In this paper, we propose feature-based adversarial meta-embeddings (FAME) that (1) align the embedding spaces with adversarial training, and (2) use attention for combining embeddings with a layer that is guided by features reflecting word-specific properties, such as the shape or frequency of the word and, thus, can help the model to assess the quality of the different embeddings. By using attention, we avoid the shortcomings of concatenation (high-dimensional input vectors) and averaging (merging information without focus). Further, our contributions mitigate the challenges of previous attention-based meta-embeddings:
In our analysis, we show that the first contribution is especially beneficial when embeddings of different \sizes are combined. The second helps, in particular, when combining word-based with \subword-based embeddings.

We conduct experiments across a variety of tasks, languages and domains, including sequence-labeling tasks (named entity recognition (NER) for four languages, concept extraction for two special domains (clinical and materials science), and part-of-speech tagging (POS) for 27 languages) and sentence classification tasks (question classification in different domains).
Our results and analyses show that FAME outperforms existing meta-embedding methods and that even powerful \finetuned transformer models can benefit from additional embeddings using our method. 
In particular, FAME sets the new state of the art for POS tagging in all 27 languages, for NER in two languages, as well as on all tested concept extraction and two question classification datasets.

In summary, our contributions are meta-embeddings with (i) adversarial training and (ii)~a feature-based attention function. 
(iii) We perform broad experiments, ablation studies and analyses which demonstrate that our method is highly effective across tasks, domains and languages, including low-resource settings.
(iv) Moreover, we show that even representations from large-scale \pretrained transformer models can benefit from our meta-embeddings approach.
The code for FAME is publicly available\footnote{\url{https://github.com/boschresearch/adversarial_meta_embeddings}} and compatible with the flair framework \cite{flair/Akbik18}.

%% file: sections/related_work.tex
\section{Related Work}\label{sec:related}
This section surveys related work on meta-embeddings, attention and adversarial training.

\textbf{Meta-Embeddings.}
\label{sec:relatedWorkEmbeddings}
Previous work has seen performance gains by, for example, combining various types of word embeddings \cite{tsuboi-2014-neural} or 
the same type
trained on different corpora \cite{luo2014}.
For the combination, some alternatives have been proposed, such as different input channels of a convolutional neural network \cite{kim-2014-convolutional,zhang-etal-2016-mgnc}, concatenation followed by dimensionality reduction \cite{yin-schutze-2016-learning} or averaging of embeddings \cite{coates-bollegala-2018-frustratingly}, e.g., for combining embeddings from multiple languages
~\cite{lange-etal-2020-choice,reid2020combining}.
More recently, auto-encoders \cite{bollegala-bao-2018-learning,wu-etal-2020-task}, ensembles of sentence encoders \cite{poerner-etal-2020-sentence}
and attention-based methods \cite{kiela-etal-2018-dynamic,lange-etal-2019-nlnde-pharmaconer} have been introduced. The latter allows a dynamic (input-based) combination of multiple embeddings.
\newcite{emb/Winata19} and \newcite{Priyadharshini2020} used similar attention functions to combine embeddings from different languages for NER in code-switching settings. \newcite{LIU2021410} explored the inclusion of domain-specific semantic structures to improve meta-embeddings in non-standard domains. 
In this paper, we follow the idea of attention-based meta-embeddings and propose task-independent methods for improving them.

\textbf{Extended Attention.}
Attention has been introduced in the context of machine translation \cite{bahdanau2014neural} and is since then widely used in NLP
\cite[i.a.,][]{tai-etal-2015-improved, xu15-show-attend-tell, yang-etal-2016-hierarchical,vaswani2017}.
Our approach extends this technique by integrating word features into the attention function. 
This is similar to extending the source of attention for uncertainty detection \cite{adel-schutze-2017-exploring} or relation extraction \cite{zhang-etal-2017-position,li-etal-2019-improving}.
However, in contrast to these works, we use task-independent features derived from the token itself. Thus, we can use the same attention function for different tasks.

\textbf{Adversarial Training.}
Further, our method is motivated by the usage of adversarial training \cite{goodfellow2014} for creating input representations that are independent of a specific domain or feature.
This is related to using adversarial training for domain adaptation \cite{ganin2016} or coping with bias or confounding variables
\cite{li-etal-2018-towards,raff2018,zhang2018bias,barrett-etal-2019-adversarial,mchardy-etal-2019-adversarial}. Following \newcite{ganin2016}, we use gradient reversal training in this paper.
Recent studies use adversarial training on the word level 
to enable cross-lingual transfer from a source to a target language \cite{zhang-etal-2017-adversarial,keung-etal-2019-adversarial,wang-etal-2019-weakly,bari2020}.
In contrast, our discriminator is not binary but multinomial (as in \newcite{chen-cardie-2018-multinomial})
and allows us to create a common space for embeddings from different \types.

%% file: sections/approach_fame.tex
\section{Meta-Embeddings}\label{sec:approach}
\begin{figure}
	\centering
	\includegraphics[width=\columnwidth]{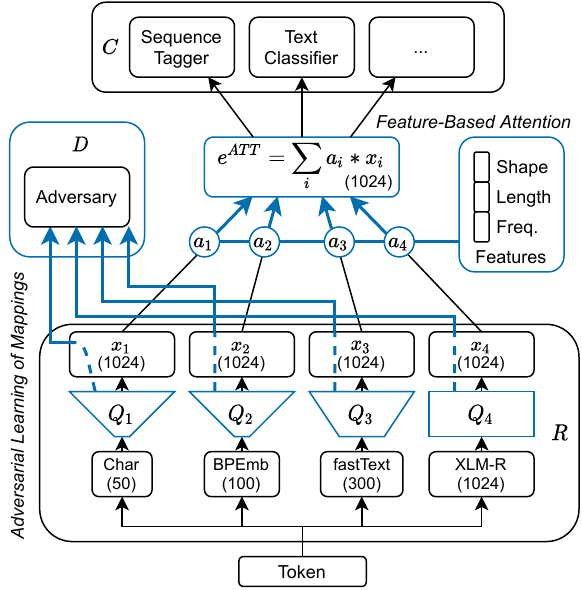}
	\caption{Overview of the FAME model architecture. Blue lines highlight our contributions. 
	$C$ (classifier), $D$ (discriminator) and $R$ (input representation) denote the components of adversarial training. The \sizes of intermediate representations are given in parentheses. }
	\label{fig:model}
\end{figure}

In this section, we present our proposed FAME model with feature-based meta-embeddings with adversarial training. The FAME model is depicted in Figure~\ref{fig:model}.

\renewcommand{\vec}[1]{\mathbf{#1}}

\subsection{Attention-Based Meta-Embeddings}
As some embeddings are more effective in modeling certain words, e.g., domain-specific embeddings for in-domain words, we use attention-based meta-embeddings that are able to combine different embeddings dynamically as introduced by \newcite{kiela-etal-2018-dynamic}. 

Given $n$ embeddings $e_1 \in \mathbb{R}^{E_1}$, ... $e_n \in \mathbb{R}^{E_n}$ of potentially different \sizes $E_1$, ... $E_n$, they first need to be mapped to the same space (with $E$ dimensions): $x_i = \tanh(Q_i \cdot e_i + b_i), 1 \le i \le n$.
Note that the mapping parameters $Q_i \in \mathbb{R}^{E \times E_i}$ and $b_i \in \mathbb{R}^{E}$ are learned for each embedding method during training of the downstream task.
Then, attention weights $\alpha_i$ are computed by:
\begin{equation}
\alpha_i = \frac{\exp(V \cdot \tanh(W x_i))}{\sum_{l=1}^n \exp(V \cdot \tanh(W x_l))}
\label{eq:attBaseline}
\end{equation} 
with $W \in \mathbb{R}^{H \times E}$ and $V \in \mathbb{R}^{1 \times H}$ being parameter matrices that are randomly initialized and learned during training.
Finally, the embeddings $x_i$ are weighted using the attention weights $\alpha_i$ resulting in the word representation: 
\begin{equation}
e^{ATT} = \sum_i \alpha_i \cdot x_i
\end{equation}
This approach requires the model to learn parameters for the mapping function as well as for the attention function. The first might be challenging if the original embeddings have different \sizes while the latter might be hard if the embeddings represent inputs from different \types, such as words vs.\ \subwords. We support this claim experimentally in our analysis in Section \ref{sec:types}.

\subsection{Feature-Based Attention} \label{sec:feature-based-attention}

Equation \ref{eq:attBaseline} for calculating attention weights only depends on $x_i$, the representation of the current word.\footnote{\newcite{kiela-etal-2018-dynamic} proposed two versions: using the word embeddings or using the hidden states of a bidirectional LSTM encoder. Our observation holds for both of them.}
While this can be enough when only standard word embeddings are used, \subword- and character-based embeddings are able to create vectors for out-of-vocabulary inputs and distinguishing these from tailored vectors for frequent words is challenging without further information (see Section~\ref{sec:types}). 
To allow the model to make an informed decision which embeddings to focus on, we propose to use the features described below as an additional input to the attention function. The word features are represented as a vector $f \in \mathbb{R}^F$ and integrated into the attention function (Equation \ref{eq:attBaseline}) as follows:
\begin{align}
	\alpha_i = \frac{\exp(V \cdot \tanh(W x_i + U f))}{\sum_{l=1}^n \exp(V \cdot \tanh(W x_l + U f))}
	\label{eq:attention}
\end{align}
with $U \in \mathbb{R}^{H \times F}$ being a parameter matrix that is learned during training.

\paragraph{Features.}
FAME uses the following task-independent features based on word characteristics. 

\emph{- Length}: 
Long words, in particular compounds, 
are often less frequent in embedding vocabularies, 
such that the word length can be an indicator for rare or out-of-vocabulary words. 
We encode the lengths in 20-dimensional one-hot vectors. Words with more than 19 characters share the same vector.

\emph{- Frequency}: 
High-frequency words can typically be modeled well by word-based embeddings, while low-frequency words are better captured with \subword-based embeddings.
Moreover, frequency is domain-dependent and can thus help to decide between embeddings from different domains.
We estimate the frequency $n$ of a word in the general domain from its rank $r$ in the \fastText-based embeddings provided by \newcite{fastText/Grave18}:
$n(r) = k / r$ with $k = 0.1$, following \newcite{manning1999foundations}.
Finally, we group the words into 20 bins as in~\newcite{mikolov12-lm} and represent their frequency with a 20-dimensional one-hot vector.

\emph{- Word Shape}: 
Word shapes capture certain linguistic features 
and are often part of manually designed feature sets, e.g., for CRF classifiers~\cite{lafferty01-crf}. 
For example, uncommon word shapes can be indicators for domain-specific words, which can benefit from domain-specific embeddings.
We create 12 binary features that capture information on the word shape, including whether the first, any or all characters are uppercased, alphanumerical, digits or punctuation marks.

\emph{- Word Shape Embeddings:}
In addition, we train word shape embeddings (25 dimensions) similar to \newcite{limsopatham-collier-2016-bidirectional}. 
For this, the shape of each word is converted by replacing letters with \textit{c} or \textit{C} (depending on the capitalization), digits with \textit{n} and punctuation marks with \textit{p}. For instance, \textit{Dec. 12th} would be converted to \textit{Cccp nncc}.
The resulting shapes are one-hot encoded and a trainable randomly initialized linear layer is used to compute the shape representation.

All sparse feature vectors (binary or one-hot encoded) are fed through a linear layer to generate a dense representation.
Finally, all features are concatenated into a single feature vector $f$ of 77 dimensions which is used in the attention function as described earlier.

\subsection{Adversarial Learning of Mappings}
\label{sec:adversarial}
The attention-based meta-embeddings require that all embeddings have the same \size for summation. 
For this, mapping matrices need to be learned, as only a limited number of embeddings exist for many languages and domains, and there is typically no option to only use embeddings of the same size. 
To learn an effective mapping, we propose to use adversarial training.
In particular, FAME adapts gradient-reversal training with three components:
the representation module $R$ consisting of the different embedding models and the mapping functions $Q$ to the common embedding space, a discriminator $D$ that tries to distinguish the different embeddings from each other, and a downstream classifier $C$ which is either a sequence tagger or a sentence classifier in our experiments (and is described in more detail in Section~\ref{sec:architectures}).

The input representation is shared between the discriminator and downstream classifier and trained with gradient reversal to fool the discriminator.
To be more specific, the discriminator $D$ is a multinomial non-linear classification model with a standard cross-entropy loss function $L_D$.
In our sequence tagging experiments, the downstream classifier $C$ has a conditional random field (CRF) output layer and is trained with a CRF loss $L_C$ to maximize the log probability of the correct tag sequence \cite{ner/Lample16}. In our sentence classification experiments, $C$ is a multinomial classifier with cross-entropy loss $L_C$.
Let $\theta_R$, $\theta_D$, $\theta_C$ be the parameters of the representation module, discriminator and downstream classifier, respectively.
Gradient reversal training will update the parameters as follows:
\begin{align}
	\theta_D &= \theta_D - \eta \lambda \frac{\partial L_D}{\partial \theta_D}; \;\;\;
	\theta_C = \theta_C - \eta \frac{\partial L_C}{\partial \theta_C}\\
	\theta_R &= \theta_R - \eta (\frac{\partial L_C}{\partial \theta_R} - \lambda \frac{\partial L_D}{\partial \theta_R})
\end{align}
with $\eta$ being the learning rate and $\lambda$ being a hyperparameter to control the discriminator influence.

\section{Neural Architectures}
\label{sec:architectures}
In this section, we present the architectures we use for text classification and sequence tagging. 
Note that our contribution concerns the input representation layer, which can be used with any NLP model, e.g., also sequence-to-sequence models. 

\begin{table}
\footnotesize
\centering
\begin{tabular}{lll} \toprule
 & Dimensions & \begin{tabular}[c]{@{}l@{}}\Finetuned?\\
 \end{tabular} \\ \midrule
\multicolumn{3}{l}{\textit{General Domain}} \\
Character & 50 & Yes \\
BPEmb & 100 & No \\
\fastText & 300 & No \\
XLM-R & 1024 & No / Yes \\ \midrule
\multicolumn{3}{l}{\textit{Domain-specific}} \\
Word & 100 (En), 300 (Es) & No \\
Transformer & 768 (En) & No / Yes \\ \bottomrule
\end{tabular}
\caption{Overview of embeddings used in our models. }
\label{tab:embeddings}
\end{table}

\subsection{Input Layer}\label{sec:embeddings}
The input to our neural networks is our FAME meta-embeddings layer as described in Section \ref{sec:approach}. 
Our methodology does not depend on the embedding method, i.e., it can incorporate any token representation.
In our experiments, we use the embeddings listed in Table~\ref{tab:embeddings} based on insights from related work. 
In particular, \newcite{flair/Akbik18} showed the advantages of character and \fastText embeddings \cite{fastText/bojanowski2017} and
\newcite{bpemb/heinzerling18} showed similar results for character and BPE embeddings. Thus, we decided to use the union (char+\fastText+BPE) with a state-of-the-art multilingual Transformer \cite[][XLM-R]{conneau2019unsupervised}.
Our character-based embeddings are randomly initialized and accumulated to token embeddings using a bidirectional long short-term memory network \cite{lstm/Hochreiter97} with 25 hidden units in each direction. 

For experiments in non-standard domains, we add domain-specific embeddings, including word embeddings from the clinical domain for English \cite{emb/bio/Moen13} and Spanish \cite{gutierrezfandino2021spanish} and the materials science domain \cite{tshitoyan2019unsupervised}. Further, we include domain-specific transformer models for experiments on English data, i.e.,
Clinical BERT \cite{alsentzer-etal-2019-publicly} trained on MIMIC,
and SciBERT \cite{beltagy-etal-2019-scibert} trained on academic publications from semantic scholar. 

For all experiments, our baselines and proposed models use the same set of embeddings.
We experiment with both freezing and \finetuning the transformer embeddings during training. However, note that \finetuning the transformer model increases the model size by more than a factor of 100 from 4M trainable parameters to 535M as shown in Table~\ref{tab:parameters}. 
This increases computational costs by a large margin. 
For example, in our experiments, the time for training a single epoch for English NER increases from 3 to 38 minutes.

\begin{table}
\setlength\tabcolsep{4.8pt}
\footnotesize
\centering
\begin{tabular}{lcc} \toprule
 & \multicolumn{2}{c}{\begin{tabular}[c]{@{}c@{}}Transformer \\ \finetuned?\end{tabular}} \\ 
Meta-embeddings method & \multicolumn{1}{c}{No} & \multicolumn{1}{c}{Yes} \\ \midrule
\multicolumn{3}{l}{\textit{General Domain (4 embeddings)}} \\
Concatenation           & 10.0 / 3.4 & 543.9 / 539.4 \\
Attention-based meta-emb & 4.0 / 4.0 & 537.9 / 538.9 \\
Feature-based attention  & 4.0 / 4.0 & 538.0 / 538.9 \\ 
\midrule
\multicolumn{3}{l}{\textit{Domain-specific (4+2 embeddings)}} \\
Concatenation           & 14.9 / 5.3 & 652.2 / 648.2 \\
Attention-based meta-emb & 4.9 / 4.9 & 642.2 / 643.2 \\
Feature-based attention  & 5.0 / 4.9 & 642.2 / 643.2 \\ 
\midrule
+ Adversarial Discriminator & +1.0 / +1.0 & +1.0 / +1.0\\
\bottomrule
\end{tabular}
\caption{Number of trainable parameters (in million) of our models for sequence labeling / text classification.}
\label{tab:parameters}
\setlength\tabcolsep{6pt}
\end{table}

\subsection{Model for Sequence Tagging}
\label{sec:system}
Our sequence tagger follows a well-known architecture \cite{ner/Lample16} with a
 bidirectional long short-term memory (BiLSTM)  network and conditional random field (CRF) output layer \cite{lafferty01-crf}.
Note that we perform sequence tagging on sentence level without cross-sentence context as done, i.a., by \newcite{schweter2020flert}.

\subsection{Models for Text Classification}
For sentence classification, we use a BiLSTM sentence encoder. 
The resulting sentence representation is fed into a linear layer followed by a softmax activation that outputs label probabilities.

\subsection{Hyperparameters and Training}
To ensure reproducibility, we describe details of our models and training procedure in the following. 

\paragraph{Hyperparameters.}
We use hidden sizes of 256 units per direction for all BiLSTMs. 
The attention layer has a hidden size $H$ of 10. 
We set the mapping size $E$ to the size of the largest embedding in all experiments, i.e., 1024 dimensions, the size of XLM-R embeddings. 
The discriminator $D$ has a hidden size of 1024 units and is trained every 10$^{th}$ batch.
We perform a hyperparameter search for the $\lambda$ parameter in \{1e-4, 1e-5, 1e-6, 1e-7\} for models using adversarial training. 
Note that we use the same hyperparameters for all models and all tasks.

\paragraph{Training.}
We use the AdamW optimizer with an initial learning rate of 5e-6. 
We train the models for a maximum of 100 epochs and select the best model according to the performance using the task's metric on the development set if available, or using the training loss otherwise. 
The training was performed on Nvidia Tesla V100 GPUs with 32GB VRAM.\footnote{All experiments ran on a carbon-neutral GPU cluster. }

%% file: sections/experiments_fame.tex
\section{Experiments and Results}\label{sec:experiments}
We now describe the tasks and datasets 
we use in our experiments as well as our results. 

\begin{table*}
	\centering 
	\footnotesize
	\setlength\tabcolsep{5pt}
	\begin{tabular}{l|cccc}	\toprule
		& \multicolumn{4}{c}{NER} \\
		Model & En & De & Es & Nl \\ \midrule
		\newcite{schweter2020flert} & 93.69 & \textbf{92.29} & 89.93 & 94.66 \\
		\newcite{yu2020named} & 93.5 & 90.3 & \textbf{90.3} & 93.7 \\
		XLM-R \cite{conneau2019unsupervised} & 92.92 & 85.81 & 89.72 & 92.53 \\
		FAME (our model) & \textbf{94.11} & 92.28 & 89.90 & \textbf{95.42} \\ 
		\bottomrule
	\end{tabular}
	\hfill
	\begin{tabular}{l|ccc} \toprule
		& \multicolumn{3}{c}{Concept Extraction} \\
		Model & Clin$_{\mathrm{En}}$ & Clin$_{\mathrm{Es}}$ & Sofc$_{\mathrm{En}}$ \\ \midrule
		\newcite{alsentzer-etal-2019-publicly} & 87.7 & - & - \\
		\newcite{lange2020closing} & 88.9 & 91.4 & -- \\
		\newcite{friedrich-etal-2020-sofc} & -- & -- & 81.5\\
		FAME (our model) & \textbf{90.08} & \textbf{92.68} & \textbf{83.68} \\
		\bottomrule
	\end{tabular}
	\caption{NER and concept extraction results (\fscore). XLM-R is a \finetuned transformer \cite{conneau2019unsupervised}. } 
	\setlength\tabcolsep{6pt}
	\label{tab:sl_results}
\end{table*}

\subsection{Tasks and Datasets}
\paragraph{Sequence Labeling.}
For sequence labeling, we use named entity recognition (NER) and part-of-speech tagging (POS) datasets from different domains and languages. 
For NER, we use the CoNLL benchmark datasets from the news domain (English/German/Dutch/Spanish) \cite{data/conll/Sang02,data/conll/Sang03}.
In addition,
we conduct experiments for concept extraction on two datasets from the clinical domain, the English i2b2 2010 data \cite{i2b2/task/uzuner2010} and the Spanish PharmaCoNER task \cite{gonzalez-agirre-etal-2019-pharmaconer}, as well as experiments on the materials science domain \cite{friedrich-etal-2020-sofc}. 
For POS tagging, we use the universal dependencies treebanks version 1.2 (UPOS tag) and use the 27 languages for which \newcite{yasunaga-etal-2018-robust} reported numbers.

\paragraph{Sentence Classification.}
We experiment with three question classifications tasks, namely the TREC corpus \cite{voorhees1999trec} with 6 or 50 labels and GARD \cite[][clinical domain]{kilicoglu-etal-2016-annotating}.

\subsection{Evaluation Results}\label{sub:results}
We now present the results of our experiments. All reported numbers are the averages of three runs.

\paragraph{Sequence Labeling.}
Tables \ref{tab:sl_results} and \ref{tab:pos_results} show the results for sequence labeling in comparison to the state of the art.\footnote{Following prior work, we report the micro-\fscore for the NER and clinical corpora, the macro-\fscore for the SOFC corpus and accuracy for the POS corpora.}
Our models consistently set the new state of the art for English and Dutch NER, for domain-specific concept extraction as well as for all 27 languages for POS tagging. 
The comparison with XML-R on NER shows that our FAME method can also improve upon already powerful transformer representations.
In domain-specific concept extraction, we outperform related work by 1.5 \fscore-points on average.
This shows that our approach works across languages and domains.

\newcommand{\dashedrule}{\midrule}
\begin{table}[t!]
	\centering
	\footnotesize
	\setlength\tabcolsep{5.5pt}
	\begin{tabular}{l|ccc|c}
		\toprule
		& \multicolumn{1}{c}{SOTA1} 
		& \multicolumn{1}{c}{SOTA2}
		& \multicolumn{1}{c|}{SOTA3}
		& \multicolumn{1}{c}{FAME} \\
		\midrule
		Bg (Bulgarian)  & 97.97 & 98.53 & 98.7 & \textbf{99.53} \\
		Cs (Czech)      & 98.24 & 98.81 & 98.9 & \textbf{99.33} \\
		Da (Danish)     & 96.35 & 96.74 & 97.0 & \textbf{99.13} \\
		De (German)     & 93.38 & 94.35 & 94.0 & \textbf{95.95} \\
		En (English)    & 95.17 & 95.82 & 95.6 & \textbf{98.09} \\
		Es (Spanish)    & 95.74 & 96.44 & 96.5 & \textbf{97.75} \\
		Eu (Basque)     & 95.51 & 94.71 & 95.6 & \textbf{97.66} \\
		Fa (Persian)    & 97.49 & 97.51 & 97.1 & \textbf{98.68} \\
		Fi (Finnish)    & 95.85 & 95.40 & 94.6 & \textbf{98.67} \\
		Fr (French)     & 96.11 & 96.63 & 96.2 & \textbf{97.19} \\
		He (Hebrew)     & 96.96 & 97.43 & 96.6 & \textbf{98.00} \\
		Hi (Hindi)      & 97.10 & 97.21 & 97.0 & \textbf{98.35} \\
		Hr (Croatian)   & 96.82 & 96.32 & 96.8 & \textbf{97.96} \\
		Id (Indonesian) & 93.41 & 94.03 & 93.4 & \textbf{94.24} \\
		It (Italian)    & 97.95 & 98.08 & 98.1 & \textbf{98.82} \\
		Nl (Dutch)      & 93.30 & 93.09 & 93.8 & \textbf{94.74} \\
		No (Norwegian)  & 98.03 & 98.08 & 98.1 & \textbf{99.16} \\
		Pl (Polish)     & 97.62 & 97.57 & 97.5 & \textbf{99.05} \\
		Pt (Portuguese) & 97.90 & 98.07 & 98.2 & \textbf{98.86} \\
		Sl (Slovenian)  & 96.84 & 98.11 & 98.0 & \textbf{99.44} \\
		Sv (Swedish)    & 96.69 & 96.70 & 97.3 & \textbf{99.17} \\
		\cline{1-5}
		Avg.            & 96.40 & 96.65 & 96.6 & \textbf{98.08} \\
		\midrule
		El (Greek)      & - & 98.24 & 97.9 & \textbf{98.89} \\
		Et (Estonian)   & - & 91.32 & 92.8 & \textbf{97.07} \\
		Ga (Irish)      & - & 91.11 & 91.0 & \textbf{94.27} \\
		Hu (Hungarian)  & - & 94.02 & 94.0 & \textbf{97.72} \\
		Ro (Romanian)   & - & 91.46 & 89.7 & \textbf{96.64} \\
		Ta (Tamil)      & - & 83.16 & 88.7 & \textbf{91.10} \\
		\cline{1-5}
		Avg.            & - & 91.55 & 92.4 & \textbf{95.95} \\
		\bottomrule
	\end{tabular}
    \caption{POS tagging results (accuracy)  (using gold segmentation). SOTA1 refers to results from \newcite{plank-etal-2016-multilingual}, SOTA2 to \newcite{yasunaga-etal-2018-robust} and SOTA3 to \newcite{heinzerling-strube-2019-sequence}. As \newcite{yasunaga-etal-2018-robust}, we split into high-resource (top) and low-resource languages (bottom). 
    }
    \label{tab:pos_results}
	\setlength\tabcolsep{6pt}
\end{table}

\begin{table*}
	\centering 
	\footnotesize
	\setlength\tabcolsep{4.3pt}
	\begin{tabular}{ll|cccc|ccc|ccc}
		\toprule
		& corresponding to& \multicolumn{4}{c|}{NER}
		& \multicolumn{3}{c|}{Concept Extraction}
		& \multicolumn{3}{c}{POS (subset)} \\
		Model & baseline meta-emb.& En & De & Es & Nl & Clin$_{\mathrm{En}}$ & Clin$_{\mathrm{Es}}$ & Sofc$_{\mathrm{En}}$ & Et & Ga & Ta \\ \midrule
		\multicolumn{2}{l|}{FAME (our model, w/ \finetuning)}  & 94.11 & 92.28 & 89.90 & 95.42 & 90.08 & 92.68 & 83.68 & 97.07 & 94.27 & 91.10 \\
		\midrule
		\multicolumn{2}{l|}{FAME (our model, w/o \finetuning)} & 93.43 & \underline{91.96} & \underline{88.86} & 93.28 & 89.23 & \underline{91.97} & 81.85 & \underline{96.03} & \underline{91.47} & \underline{89.58} \\
		-- features     &               & 93.37 & 91.66 & 88.37 & 92.98 & 89.07 & 91.42 & 81.48 & 95.81 & 90.20 & 88.73 \\
		-- adversarial  & Attention {\scriptsize(ATT)}     & 93.22 & 91.52 & 88.16 & 92.46 & 88.87 & 91.33 & 81.31 & 95.19 & 87.79 & 87.93 \\
		-- attention    & Average {\scriptsize(AVG)}      & 92.38 & 90.14 & 88.44 & 92.37 & 88.69 & 90.23 & 80.28 & 93.20 & 86.95 & 87.73 \\
		-- sum, mapping & Concatenation {\scriptsize(CAT)} & 91.00 & 90.54 & 85.40 & 88.51 & 87.97 & 90.66 & 80.08 & 91.63 & 86.32 & 84.51 \\
		\bottomrule
	\end{tabular}
	\caption{Ablation study results for sequence labeling. 
	We underline our FAME models without \finetuning for which we found statistically significant differences to the attention-based meta-embeddings (ATT). } 
	\setlength\tabcolsep{6pt}
	\label{tab:ablation_sl}
\end{table*}

\begin{table}
	\centering 
	\footnotesize
	\setlength\tabcolsep{4.8pt}
	\begin{tabular}{l|ccc}
		\toprule
		Model & TREC-6 & TREC-50 & GARD \\ \midrule
		\newcite{xu-etal-2020-multi} & 96.2 & 92.0 & 84.9 \\ 
		\newcite{roberts2014automatically} & - & - & 80.4 \\
		\newcite{XIA201820} & 98.0 & - & - \\
		FAME (our model) & \textbf{98.2} & \textbf{93.0} & \textbf{87.90} \\
		\bottomrule
	\end{tabular}
	\caption{Sentence classification results (accuracy). } 
	\setlength\tabcolsep{6pt}
	\label{tab:text_results}
\end{table}

\paragraph{Sentence Classification.}
Similar to sequence labeling, our FAME approach outperforms the existing machine learning models on all three tested sentence classification datasets as shown in Table~\ref{tab:text_results}. This demonstrates that our approach is generally applicable and can be used for different tasks beyond the token level.\footnote{Note that a rule-based system \cite{tayyar-madabushi-lee-2016-high} achieves 97.2\% accuracy on TREC-50. However, this requires high manual effort tailored towards this dataset and is not directly comparable to learning-based systems.}

%% file: sections/analysis_fame.tex
\section{Analysis}\label{sec:analysis}
We finally analyze the different components of our proposed FAME model by investigating, i.a., ablation studies, attention weights and low-resource settings.

\subsection{Ablation Study on Model Components}
Table \ref{tab:ablation_sl} provides an ablation study on the different components of our FAME model for exemplary sequence-labeling tasks. 

First, we ablate the \finetuning of the embedding models as we found that this has a large impact on the number of parameters of our models (538M vs.\ 4M) and, as a result, on the training time (cf., Section \ref{sec:embeddings}). 
Our results show that \finetuning does have a positive impact on the performance of our models but our approach still works very well with frozen embeddings.
In particular, our non-finetuned FAME model is competitive to a finetuned XLM-R model (see Table~\ref{tab:sl_results}) and outperforms it on 3 out of 4 languages for NER. 

Second, we ablate our two newly introduced components (features and adversarial training) and find that both of them have a positive impact on the performance of our models across tasks, languages and domains.

With successively removing components, we obtain models that actually correspond to baseline meta-embeddings as shown in the second column of the table. 
Our method without features and adversarial training, for example, corresponds to the baseline attention-based meta-embedding approach (ATT). Further removing the attention function yields average-based meta-embeddings (AVG). Finally, we also evaluate another baseline meta-embedding alternative, namely concatenation (CAT).
Note that concatenation leads to a very high-dimensional input representation and, therefore, requires more parameters in the next neural network layer, which can be inefficient in practice.

\textbf{Statistical Significance.}
To show that FAME significantly improves upon the attention-based meta-embeddings, we report statistical significance\footnote{With paired permutation testing with 2$^{20}$ permutations and a significance level of 0.05.} between those two models (using our method without \finetuning for a fair comparison).
Table \ref{tab:ablation_sl} shows that we find statistically significant differences in six out of ten settings.

\subsection{Influence of Embedding \Types and \Sizes}\label{sec:types}
Next, we perform an analysis to show the effects of our method for embeddings of different \sizes and \types and support our motivation that our contributions help in those settings.
As a testbed, we perform Spanish concept extraction and utilize the embeddings published by \newcite{fastText/Grave18} and \newcite{gutierrezfandino2021spanish} 
as they allow us to nicely isolate the desired effects. 

In particular, they
published pairs of embeddings (all having 300 dimensions) that were trained on the same corpora. The first embeddings are standard word embeddings and the second embeddings are \subword embeddings with out-of-vocabulary functionality. As both were trained on the same data, we can isolate the effect of embedding \types in a first experiment. 
In addition, \newcite{gutierrezfandino2021spanish} published smaller versions with 100 dimensions that were trained under the same conditions. We use those in a second experiment to analyze the effects of combining embeddings of different \sizes.

The results are shown in Table~\ref{tab:methods}.
We find that adversarial training becomes particularly important whenever differently-sized embeddings are combined, i.e., when the model needs to learn mappings to higher \sizes. 

Further, we see that the inclusion of our proposed features helps substantially in the presence of \subword embeddings.
The reason might be that with sets of both word-based and \subword-based embeddings, it 
gets harder to decide which embeddings are useful (e.g., word-based embeddings for high-frequency words) and should, thus, get higher attention weights. Our features have been designed in a way to explicitly guide the attention function in those cases, e.g., by indicating the frequency of a word. 
In addition, Table~\ref{tab:features} shows an ablation study on our different features for this testbed setting. We see that length and shape are the most important features, as excluding either of them reduces performance the most.

\newcommand{\dt}[1]{\scriptsize ({\tiny +}#1)}

\begin{table}
\setlength\tabcolsep{3.1pt}
\centering
\footnotesize
\begin{tabular}{l|ll|ll} \toprule
\sizesA & \multicolumn{2}{c|}{Same} & \multicolumn{2}{c}{Different} \\ 
\typesA & Word & \Subword & Word & \Subword \\ \midrule
ATT & 89.27 & 88.00 & 88.60 & 88.16 \\
+ FEAT & 89.28 \dt{.01} & 88.62 \dt{.62} & 88.64 \dt{.04} & 88.42 \dt{.26} \\
+ ADV  & 89.34 \dt{.07} & 88.31 \dt{.31} & 89.23 \dt{.63} & 88.44 \dt{.28} \\ 
\bottomrule
\end{tabular}
\caption{Effect of our proposed methods on embeddings of different \types (word vs.\ \subword) and \sizes (same vs.\ different dim.). ATT: attention-based meta-embeddings, FEAT: feature-based attention function, ADV: adversarial training of mapping. 
We add the differences between our methods and ATT.}
\label{tab:methods}
\setlength\tabcolsep{6pt}
\end{table}

\begin{table}
\centering
\footnotesize
\begin{tabular}{lll} \toprule
Attention function & \fscore & ($\Delta$) \\ \midrule
no features  & 88.0 & \\
all features & 88.62 & (+.62) \\
\ -- shape           & 88.65 & (+.65) \\
\ -- frequency       & 88.61 & (+.61) \\
\ -- length          & 88.45 & (+.45) \\
\ -- shape embedding & 88.34 & (+.34) \\ \bottomrule
\end{tabular}
\caption{Ablation Study: Features.}
\label{tab:features}
\end{table}

\subsection{Training in Low-Resource Settings}
\begin{figure}
	\centering
	\includegraphics[width=\columnwidth]{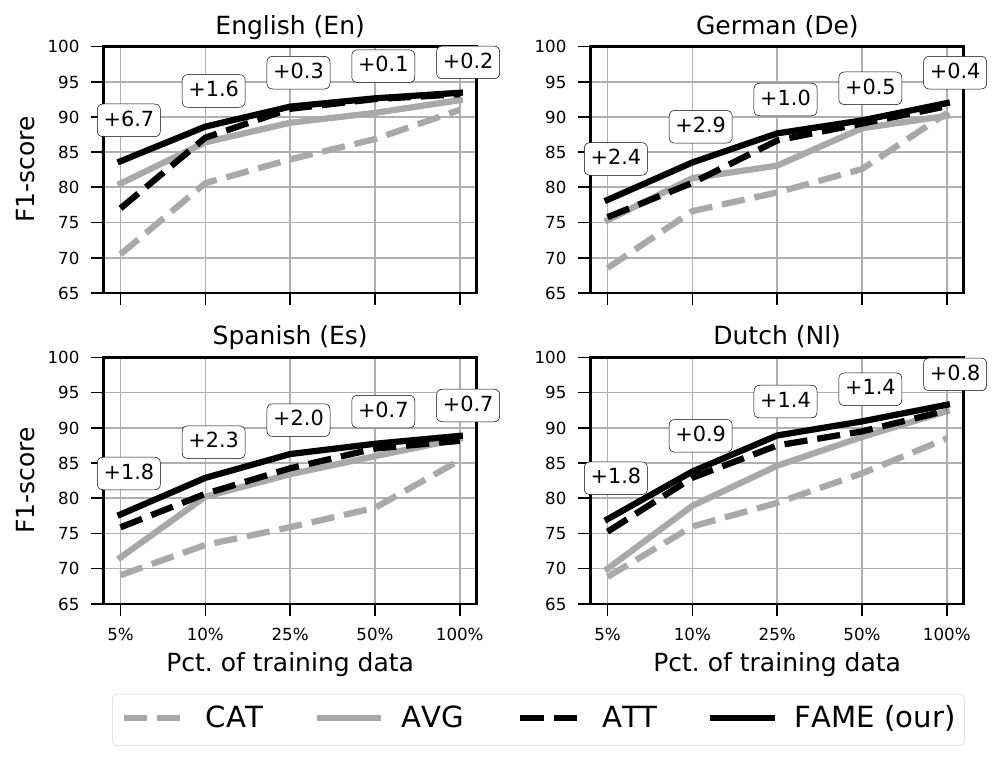}
	\caption{Performance for different training set sizes. The highlighted numbers display the difference between our FAME model without \finetuning and the attention-based meta-embeddings (ATT). Further, we compare to the baseline methods averaging (AVG) and concatenation (CAT) of embeddings.}
	\label{fig:low-res}
\end{figure}

As we observed large positive effects of our method for low-resource languages (Table \ref{tab:pos_results}), we now perform a study to further investigate this topic. We simulate low-resource scenarios by artificially limiting the training data of the CoNLL NER corpora to different percentages of all instances. The results are visualized in Figure~\ref{fig:low-res}. 
We find that the differences between the standard attention-based meta-embeddings (ATT) and our FAME method get larger with fewer training samples, with up to 6.7 \fscore points for English when 5\% of the training data is used, which corresponds to roughly 600 labeled sentences. This behavior holds for all four languages and highlights the advantages of our method when only limited training data is available. 
An interesting future research direction is the exploration of FAME for real-world low-resource domains and languages \cite{hedderich-etal-2021-survey}. 

\subsection{Analysis of Embedding Methods}\label{sub:ana_emb}

We studied the performance of each embedding method in isolation. The results are shown in Table~\ref{tab:results-embeddings} and indicate that \fastText and XLM-R embeddings are the best options in this setting. 
This observation is also reflected in the attention weights assigned by the FAME model (see Figure~\ref{fig:att_weights_feat}). In general, \fastText and XLM-R embeddings get assigned the highest weights. This highlights that the attention-based meta-embeddings are able to perform a suitable embedding selection and reduce the need for manual feature selection. 

The combination of all four embeddings is better than every single embedding, which shows the importance of combining different embeddings. In particular, the FAME model outperforms concatenation by a large margin regardless if the transformer embedding is \finetuned.

\begin{table}
\footnotesize
\centering
\begin{tabular}{lll} \toprule
 & Input Dim. & News$_{En}$ \fscore \\ \midrule
 
\multicolumn{3}{l}{\textit{Single embeddings}} \\
Character & 50   & 77.02  \\
BPEmb     & 100  & 86.37  \\
\fastText & 300  & 90.45  \\
XLM-R     & 1024 & 89.23  \\ 
\midrule

\multicolumn{3}{l}{\textit{All embeddings}} \\
CAT   & 1474 & 91.0 \\
FAME  & 1024 & 93.43 \\ 
\midrule

\multicolumn{3}{l}{\textit{\Finetuned transformer}} \\
XLM-R & 1024 & 92.12 \\
CAT   & 1474 & 92.75 \\
FAME  & 1024 & 94.11 \\
\bottomrule

\end{tabular}
\caption{Overview of embeddings used in our models. }
\label{tab:results-embeddings}
\end{table}

\subsection{Analysis of Attention Weights}

\begin{figure}[t]
	\centering
	\includegraphics[width=.49\textwidth]{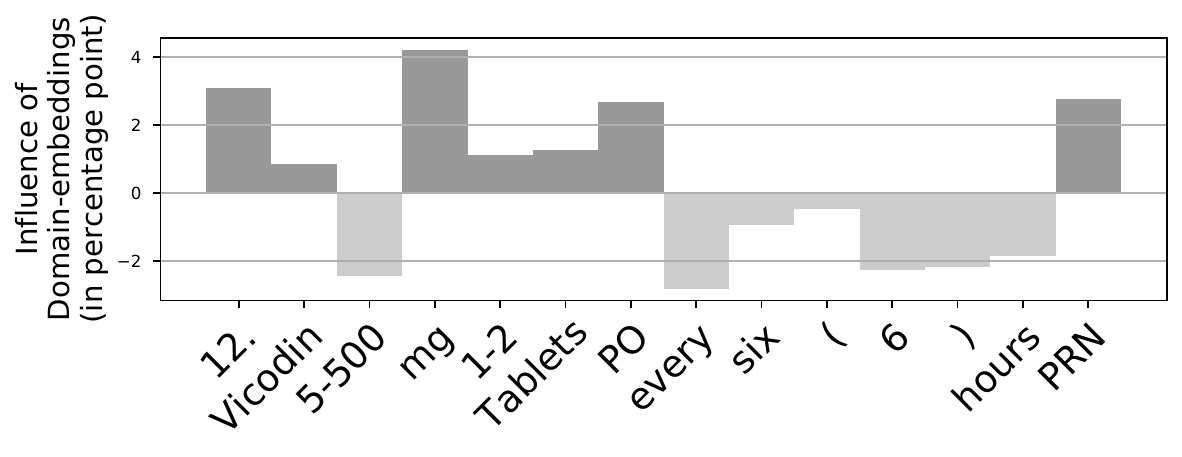}
	\vspace{-0.7cm}
	\caption{Changes in influence of domain-specific embeddings on meta-embeddings. The model prefers domain-specific embeddings for in-domain words.}
	\label{fig:att_weights_sent}
\end{figure}

Figure~\ref{fig:att_weights_sent} provides the change of attention weights from the average for the domain-specific embeddings for a sentence from the clinical domain. It shows that the attention weights for the clinical embeddings are higher for in-domain words,
such as ``mg'', ``PRN'' (which stands for ``pro re nata'') or ``PO'' (which refers to ``per os'') and lower for general-domain words, such as ``every'', ``6'' or ``hours''.
Thus, FAME is able to recognize the value of domain-specific embeddings in non-standard domains and assigns attention weights accordingly.  

\begin{figure}[t]
	\centering
	\includegraphics[width=.4\textwidth]{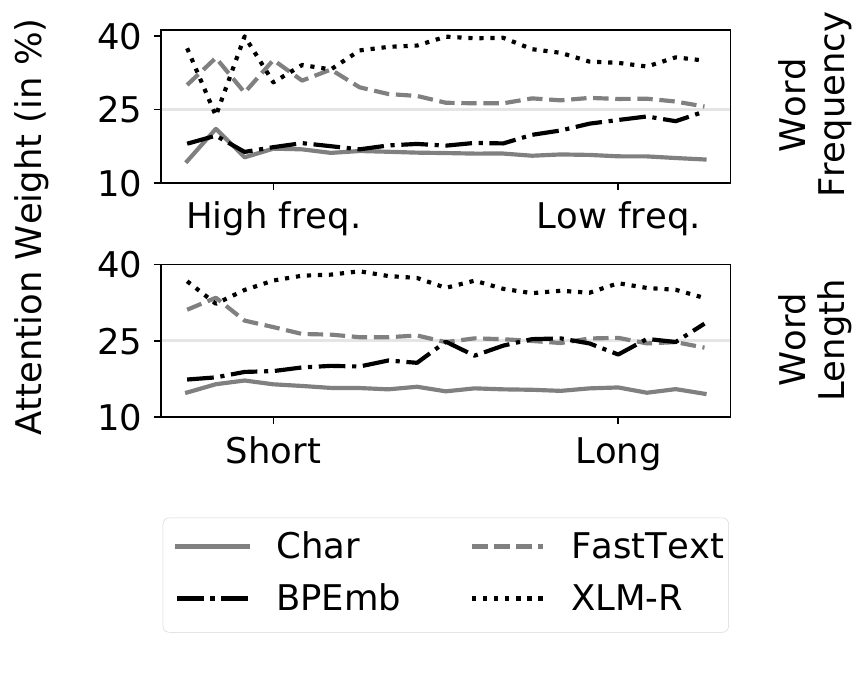}
	\caption{Attention weights assigned by the FAME model for the Clin$_{\mathrm{En}}$ corpus grouped by the features word frequency (above) and length (below).}
	\label{fig:att_weights_feat}
\end{figure}

Figure~\ref{fig:att_weights_feat} shows how attention weights change for frequency and length features as introduced in Section~\ref{sec:feature-based-attention}. In particular, it demonstrates that subword-based embeddings (BPEmb and XLM-R) get more important for long and infrequent words which are usually not well covered in the fixed vocabulary of standard word embeddings.

\subsection{Analysis of Adversarial Training}
\begin{figure}[t]
	\centering
	\includegraphics[trim=0 119 0 0,clip,width=.49\textwidth]{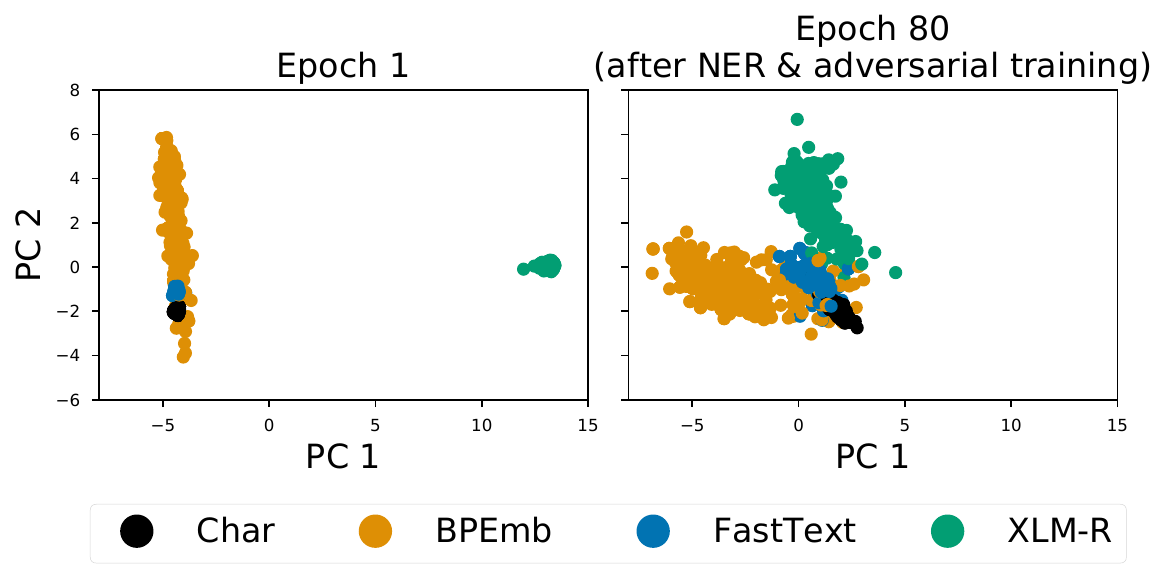}
	\caption{The meta-embeddings space before (left) and after NER and adversarial training (right).}
	\label{fig:adv_space}
\end{figure}

In contrast to previous work \cite{lange2020adversarial}, we show that adversarial training is also beneficial and boosts performance in a monolingual case when combining multiple embeddings. The embeddings were trained independent from each other. Thus, the individual embedding spaces are clearly separated. Adversarial training shifts all embeddings closer to a common space as shown in Figure~\ref{fig:adv_space}, which is important if the average is taken for the attention-based meta-embeddings approach.

%% file: sections/conclusions_fame.tex
\section{Conclusions}\label{sec:conclusion}
In this paper, we proposed feature-based adversarial meta-embeddings (FAME) to effectively combine several embeddings.
The features are designed to guide the attention layer when computing the attention weights, in particular for embeddings representing different input \types, such as \subwords or words.
Adversarial training helps to learn better mappings when embeddings of different \sizes are combined.
We demonstrate the effectiveness of our approach on a variety of sentence classification and sequence tagging tasks across languages and domains and set the new state of the art for POS tagging in 27 languages, for domain-specific concept extraction on three datasets, for NER in two languages, as well as on two question classification datasets.
Further, our analysis shows that our approach is particularly successful in low-resource settings.
A future direction is the evaluation of our method on sequence-to-sequence tasks or document representations.